\documentclass[aps,prx,twocolumn,nofootinbib]{revtex4-2}
\usepackage{graphicx}
\usepackage{mathrsfs}
\usepackage{color}
\graphicspath{{figures/}{fig/}}
\usepackage{amsmath}
\usepackage{amssymb}
\usepackage{bm}
\usepackage{slashed}
\usepackage{epsfig}
\usepackage{enumitem}
\usepackage{amsfonts}
\usepackage{epstopdf}
\usepackage{hyperref}
\usepackage{bbm}
\usepackage{textcomp}
\usepackage{color}
\usepackage[dvipsnames]{xcolor}
\usepackage{xparse}  % \NewDocumentCommand
\usepackage{booktabs}
\usepackage{threeparttable}
\usepackage[ruled,linesnumbered]{algorithm2e}
\newcommand{\sect}[1]{\section{#1}}

\allowdisplaybreaks[3]

\NewDocumentCommand{\commentModify}{m o m}{%
    \textcolor{red}{\sout{#1}}
    \IfValueT{#2}{
        \textcolor{cyan}{(\textbf{COMMENT}: \uline{#2})}
    }%
    \textcolor{blue}{#3}
}

\setlist[description]{leftmargin=*,labelindent=\parindent}
\begin{document}

\title{Mastering Olympiad-Level Physics with Artificial Intelligence}

\author{Dong-Shan Jian}
\email{dsjian@stu.pku.edu.cn}
\affiliation{School of Physics, Peking University, Beijing 100871, China}

\author{Xiang Li}
\email{lix-PHY@pku.edu.cn}
\affiliation{School of Physics, Peking University, Beijing 100871, China}

\author{Chen-Xu Yan}
\affiliation{School of Physics, Peking University, Beijing 100871, China}

\author{Hui-Wen Zheng}
\affiliation{School of Physics, Peking University, Beijing 100871, China}

\author{Zhi-Zhang Bian}
\affiliation{School of Physics, Peking University, Beijing 100871, China}

\author{You-Le Fang}
\affiliation{School of Physics, Peking University, Beijing 100871, China}

\author{Bing-Rui Gong}
\affiliation{School of Electronics Engineering and Computer Science, Peking University, Beijing 100871, China}

\author{Ren-Xi He}
\affiliation{School of Physics, Peking University, Beijing 100871, China}

\author{Jing-Tian Zhang}
\affiliation{School of Physics, Peking University, Beijing 100871, China}

\author{Sheng-Qi Zhang}
\affiliation{School of Physics, Peking University, Beijing 100871, China}
\affiliation{Center for High Energy Physics, Peking University, Beijing 100871, China}

\author{Yan-Qing Ma}
\email{yqma@pku.edu.cn}
\affiliation{School of Physics, Peking University, Beijing 100871, China}
\affiliation{Center for High Energy Physics, Peking University, Beijing 100871, China}

\author{Ce Meng}
\affiliation{School of Physics, Peking University, Beijing 100871, China}

\author{Ling-Shi Meng}
\affiliation{School of Physics, Peking University, Beijing 100871, China}

\date{\today}

\begin{abstract}
Olympiad-level physics problem-solving significantly challenges both humans and artificial intelligence (AI), as it requires integrating appropriate modeling, application of physical principles, and precise calculation within long reasoning processes. In this paper, we introduce LOCA (LOgical Chain Augmentation), an AI agent framework designed for complex physics reasoning. LOCA decomposes long reasoning into serialized atomic and verifiable steps, refining the solution through an augment-review loop.  
We evaluate LOCA on the 2025 Chinese Physics Olympiad (CPhO) theory examination, a rigorous testbed renowned for its depth and complexity. The framework achieves a near-perfect score of 313 out of 320 points, significantly surpassing the top human competitor and other baseline methods. Furthermore, LOCA attains a near-perfect score of 28.6 out of 30 on the IPhO 2025 examination, demonstrating its strong generalizability across different contexts. Our work points toward the development of trustworthy AI partners in both research and education.
\end{abstract}

\maketitle

%%%%%%%%%%%%%%%%%%%%%%%%%%%%%%%%%%%%%%%%%%%%%%%%%
\sect{Introduction}
\label{sec:introduction}
Complex reasoning represents a frontier where the capabilities of artificial intelligence (AI) intersect with the fundamental methodology of physics. While Large Language Models (LLMs) have achieved remarkable performance in structured domains such as coding and mathematics \citep{brown2020language,chatgpt,achiam2023gpt,anil2023palm,touvron2023llama,touvron2023llama2,liu2024deepseek,guo2025deepseek,Claude,comanici2025gemini,gpt5,gpto4,qwq32b,team2025kimi,yang2025qwen3}, their ability to solve novel, high-level physics problems remains limited. The complexity of translating natural language into abstract models, identifying applicable physical laws and executing accurate derivations makes LLMs prone to hallucinations. Furthermore, unlike coding or mathematics where solutions can be rigorously verified, logical errors in physical reasoning are difficult to detect. Consequently, recent evaluations show that despite their vast training data, current models struggle significantly when facing the depth of advanced physics\citep{he2024olympiadbench,chow2025physbench,chung2025theoretical,feng2025physics,qiu2025phybench,zhang2025abench,zhang2025physreason,xu2025ugphysics,siddique2025physicseval,carrit2025large}. Given the strict demand for rigor in scientific research, the primary objective must be to maximize the problem-solving accuracy of LLMs.

Existing approaches to enhance LLM reasoning generally fall into three categories: test-time strategies such as Chain-of-Thought (CoT)\citep{kojima2022large,wei2022chain}, Tree-of-Thought (ToT)\citep{yao2023tree} and self-refine\citep{madaan2023self,liu2024large,liang2024internal,zhang2024sciinstruct}; agent-based frameworks utilizing external tools like Plan-and-Solve\citep{wang2023plan}, ReAct\citep{yao2022react} and Multi-Agent Debate (MAD)\citep{du2023improving,liang2023encouraging}; and resource-intensive training methods like Supervised Fine-Tuning or Reinforcement Learning \citep{zhang2024sciinstruct,lu2022dynamic,lewkowycz2022solving}. While specialized efforts like Physics Supernova \citep{qiu2025physics}, Physicsminions\citep{yu2025physicsminions}, Physics Reasoner \citep{pang2025physics}, and Mixture of Refinement Agents (MoRA) \citep{jaiswal2024improving} have made significant progress in adapting these strategies for physics, the accuracy of problem-solving remains far from satisfactory. 

Specifically, hallucinations of LLMs often lead to plausible-sounding but physically unsound derivations, while the lack of explicit logical structure makes it difficult to verify the underlying logic. We argue that to serve as reliable assistants for research or education, AI systems must move beyond statistical emulation of text and towards a structured, verifiable reasoning grounded in first principles.

In response to these challenges, we introduce LOCA (LOgical Chain Augmentation), an agent framework specifically designed to enforce rigorous, step-wise logic. By decomposing complex reasoning into logical chains of serialized atomic, verifiable steps, LOCA transforms the opaque generation process into a transparent workflow. Central to our approach is an \textit{augment-review} loop that iteratively refines LLMs' own solutions---mirroring the self-correction process of a human physicist---to ensure both physical validity and mathematical accuracy.

To rigorously evaluate this approach, we choose the Chinese Physics Olympiad\citep{cpho} (CPhO) 2025 theory examination as our primary testbed. Renowned for its depth and complexity comparable to the International Physics Olympiad (IPhO)\citep{ipho}, the CPhO demands the synthesis of diverse physical principles and creative problem-solving in novel scenarios. Critically, we conduct our evaluation promptly after the examination's release, thereby eliminating the risk of data contamination and ensuring a genuine assessment of ab initio reasoning rather than reliance on memorization. 

We demonstrate that LOCA achieves a remarkable level of proficiency, scoring 313 out of 320 on the CPhO 2025 theory section. This performance is unprecedented, surpassing the score of the top-performing human gold medalist (204). Beyond the raw score, the solutions generated by LOCA are structured and highly readable, effectively acting as valid proofs of derivation. Furthermore, to verify the generalizability of our framework beyond a single competition style, we also extend our evaluation to the IPhO 2025, confirming the method's robustness across different standards of top-tier physics problems.

Our results suggest that when constrained by a physics-informed logical architecture, LLMs have the intrinsic capability to solve problems of exceptional complexity. 
This work highlights that enforcing logically rigorous structure is a promising approach to overcoming the reliability bottleneck in automated scientific reasoning, establishing a foundational step towards AI agents functioning as trustworthy partners in frontier scientific research and advanced education.

%%%%%%%%%%%%%%%%%%%%%%%%%%%%%%%%%%%%%%%%%%%%%%%%%
\begin{figure*}[t]
	\centering
 \includegraphics[width=0.8\textwidth]{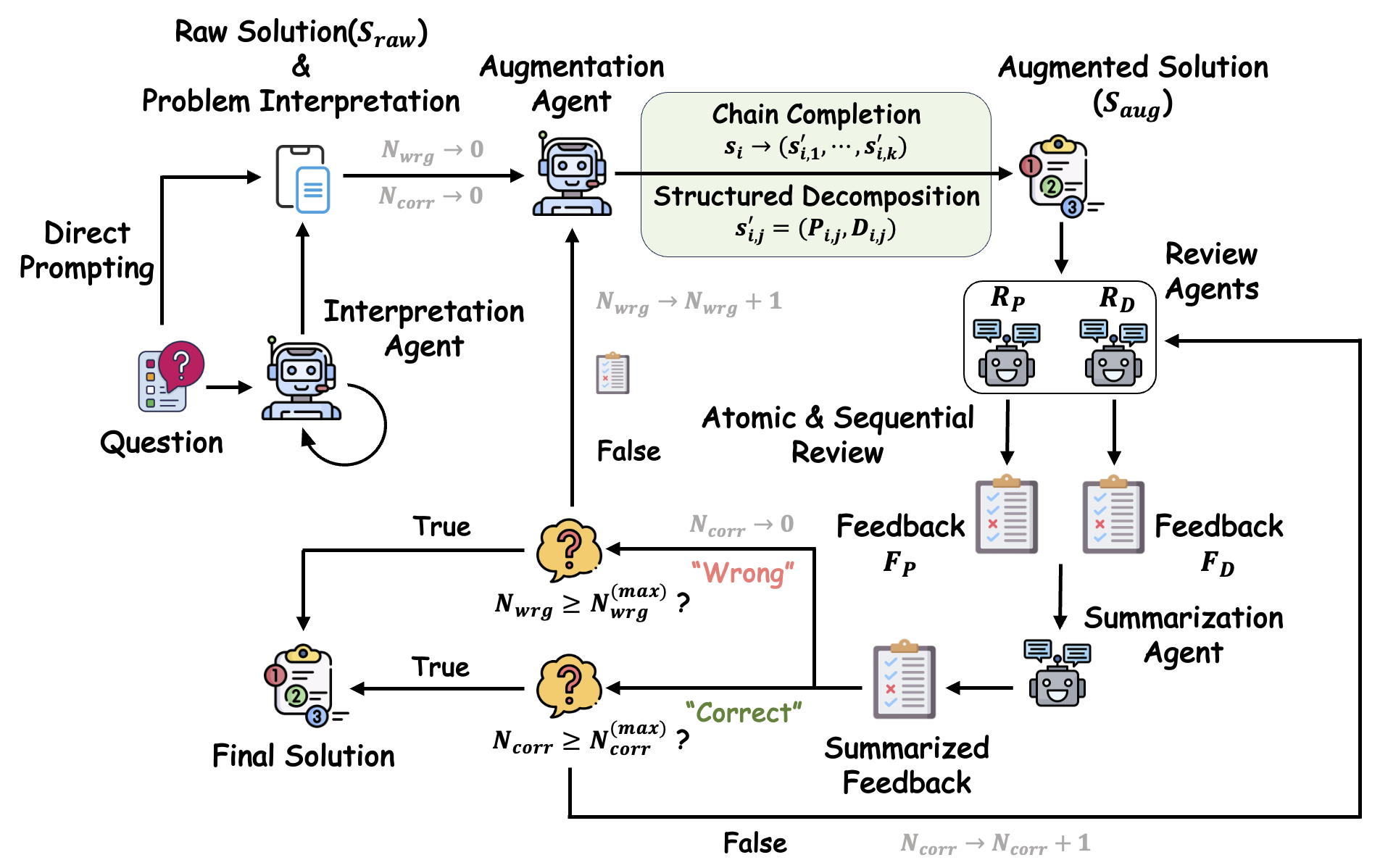} 
    \caption{\textbf{An overview of LOCA's framework.} The framework builds upon logical chain augmentation by implementing an iterative augment-and-review loop.}

	\label{fig:framework} 
\end{figure*}

\sect{Framework of LOCA}
\label{sec:loca_framework}

To address the issue of hallucinations in complex physics reasoning, LOCA decouples content generation from logical verification, employing an iterative augment-review loop to refine the solution. The framework operates through three specialized modules: (1) \textbf{Problem Interpretation}, which extracts and stabilizes the physical context; (2) \textbf{Logical Chain Augmentation}, the core component that restructures raw reasoning into atomic Principle-Derivation $(P,\,D)$ tuples; and (3) \textbf{Atomic and Sequential Review}, a mechanism that ensures the validity of each step through serialized verification. An overview of LOCA's framework is shown in Figure \ref{fig:framework}. 

While our specific implementation in test uses a single foundation model for all roles, the LOCA framework inherently allows to apply different models for drafting and reviewing. Detailed prompts for each agent role are provided in Supplemental Materials\cite{supp}.

\subsection{Problem Interpretation}

Olympiad-level problems are characterized by dense natural language with scattered symbols and subtle conditions. This complexity increases the risk of LLMs misunderstanding the initial setup, which inevitably leads to incorrect reasoning. To mitigate this, we employ a dedicated Interpretation Agent initialized in the beginning of the pipeline. Instead of attempting to solve the problem immediately, this agent is tasked solely with translating the raw problem statement ($Q_{\text{raw}}$) into a structured physical description ($Q_{\text{struct}}$). It explicitly extracts and categorizes essential information, including a canonical list of variables, specific system constraints, initial and boundary conditions, and the precise target goals. This serves as a persistent context for all subsequent steps, ensuring that the derivation remains grounded in a correct and consistent physical setup.

\subsection{Logical Chain Augmentation}
The cornerstone of LOCA is logical chain augmentation, a mechanism that transforms a raw, unstructured
solution into a detailed and structured logical chain. In standard physics reasoning, a valid derivation step consists of invoking a physical law (the principle) and applying it to the current context (the derivation). Generic LLMs, however, often conflate these aspects, producing reasoning chains where the logical basis is entangled with application. To resolve this, LOCA employs an Augmentation Agent to transform the raw solution draft $S_{\text{raw}}$ into a structural and more verifiable logical chain $S_{\text{aug}}$.

Formally, consider a raw solution draft $S_{\text{raw}} = (s_1, \dots, s_n)$, where each step $s_i$ implies a transformation of context $C_{i-1} \rightarrow C_i = C_{i-1} \cup s_i$ ($\cup$ representing the concatenation of texts). This sequence typically suffers from two key limitations: \textit{Non-Atomicity}, where multiple reasoning acts are compressed into a single step, creating opaque ``logical leap''; and \textit{Implicit Justification}, underlying physical laws or assumptions are applied without being explicitly stated. The Augmentation Agent refines $S_{\text{raw}} \to S_{\text{aug}}$ by simultaneously performing two logical operations:

\textbf{Chain Completion.}
This operation enforces atomicity by identifying and recovering reasoning steps that are often omitted in the raw draft (e.g., physical principle introduction or intermediate algebraic manipulations). Formally, a step $s_i: C_{i-1} \rightarrow C_i$ is non-atomic if it implicitly contains an intermediate context $C_{\text{int}}$ (i.e., $C_{i-1} \rightarrow C_{\text{int}} \rightarrow C_i$).  LOCA identifies and decomposes each such step into a more fundamental subsequence $(s'_{i,1}, s'_{i,2}, \dots, s'_{i,k})$ where $k > 1$.
This results in a new, more detailed solution $S_{\text{aug}} = (s'_1, s'_2, \dots, s'_m)$ with $m \ge n$, ensuring that every reasoning step represents a single, verifiable logical move rather than a composite leap.

\textbf{Structured Decomposition.}
To further distinguish the underlying general principle from its specific application, each atomic step, say $s'_j$, in the generated $S_{\text{aug}}$ is structured not as free text, but as a tuple $(P_j, D_j)$. Here, \textit{Principle} ($P_j$) is a declarative statement of the step's logical foundation. The space of valid principles extends beyond physical laws ($\mathbb{L}$) to include mathematical identities ($\mathbb{M}$) and problem-specific constraints ($\mathbb{C}$):
\begin{equation}
P_j \in \mathbb{L} \cup \mathbb{M} \cup \mathbb{C}.
\end{equation}
$P_j$ explicitly answers: \textit{``Why is this step valid?''} (e.g., ``Conservation of Momentum,'' or ``Boundary Condition at $r=R$'').
While \textit{Derivation} ($D_j$) is the specific operation required to apply $P_j$ to the preceding physical context $C_{j-1}$. This component is responsible for the explicit progress of reasoning:
\begin{equation}
D_j = \mathcal{D}(C_{j-1}, P_j), \quad C_j = C_{j-1} \cup P_j \cup D_j,
\end{equation}
where $\mathcal{D}(\cdot, P)$ denotes the application of principle $P$ to the context. $D_j$ answers: \textit{``How is the principle applied?''} (e.g., substituting variables or performing numerical calculation).

This $(P, D)$ structure is critical for precise error identification. It allows the agent to differentiate between conceptual errors and execution errors, thereby improving targeted iterative refinement. The final augmented logical chain is formalized as:
\begin{align}
S_{\text{aug}} = ((P_1, D_1), (P_2, D_2), \dots, (P_m, D_m)).
\end{align}

It is important to note that while we define Chain Completion and Structured Decomposition as distinct logical operations for clarity, the LOCA framework implements them via a unified, single-pass inference by the Augmentation Agent. The agent is prompted to directly synthesize the fully augmented $(P, D)$ sequence from the raw draft, ensuring both structural integrity and computational efficiency.

\begin{figure}[t]
	\centering
	\includegraphics[width=\columnwidth]{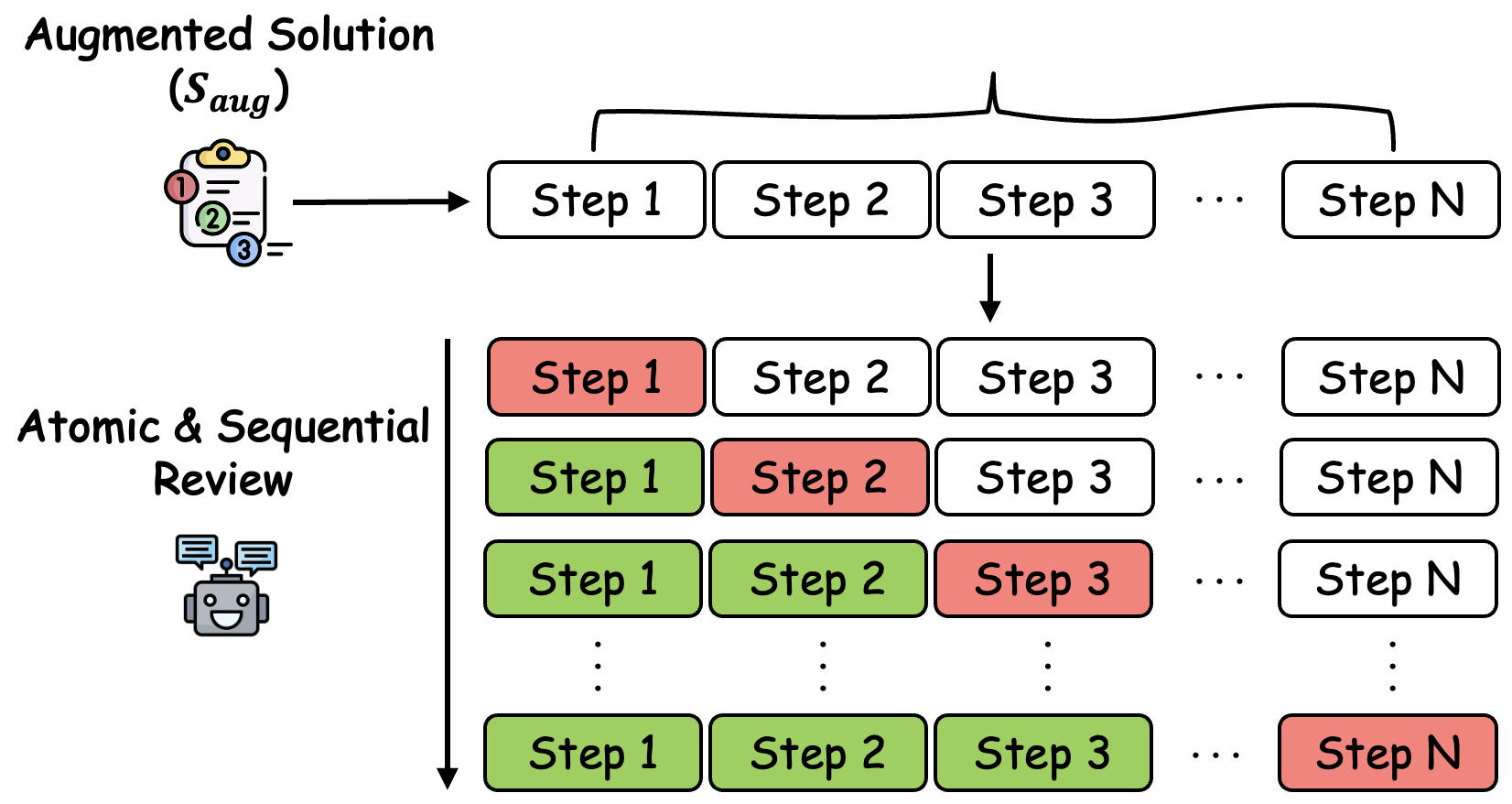} 

\caption{\textbf{The atomic and sequential review mechanism.} The mechanism sequentially iterates through each step of the solution. The step currently under review is shown in red, while the preceding steps (green), which are provisionally assumed to be correct,  form the context for the evaluation.}
	\label{fig:review}
\end{figure}

\subsection{Atomic and Sequential Review}

In rigorous scientific practice, reviewing a complex derivation is rarely a holistic judgment. A human expert does not merely glance at a whole solution to judge its correctness; rather, they trace the logical flow line-by-line, validating each step based on the immediate context. Standard LLM-based review, however, typically relies on a holistic single-pass judgment, which often overlooks subtle errors in long-chain reasoning or conflates multiple logical flaws. Mirroring the cognitive process of human verification, LOCA integrates an \textit{Atomic and Sequential Review} mechanism.

Instead of issuing a rough judgment, the Review Agent traverses the augmented solution $S_{\text{aug}} = ((P_1, D_1), \dots, (P_m, D_m))$ sequentially, as shown in Fig.~\ref{fig:review}. This hard-coded traversal forces the model to scrutinize each atomic step individually, ensuring high-resolution feedback. For each step $s_j$, the agent performs a detailed review based on the assumption that the preceding context $C_{j-1}$ is accepted. This design is crucial for error decoupling: by assuming the correctness of $C_{j-1}$ when evaluating step $s_j$, the agent focuses solely on the correctness of the current step. Notably, to ensure precision, the Review Agent consists of two specialized roles focusing on the principle ($\mathcal{R}_P$) and derivation ($\mathcal{R}_D$) respectively, where the current step is regarded as correct only if both agents agree. Furthermore, this traversal is performed concurrently in our implementation. Thus, by evaluating all steps regardless of intermediate errors, this method also ensures that a localized error in one sub-question does not compromise the review and correction of other parts.

Formally, for each step $s_j$, the review yields a correctness flag $v_j \in \{1, 0\}$ (Correct/Wrong) and specific textual feedback $f_j$:
\begin{align}
  (v_j, f_j) = \mathcal{R}(s_j | C_{j-1}) \quad \text{for } j=1, \dots, m,
\end{align}
where $f_j = \emptyset$ if $v_j=1$. Once the sequential traversal is complete, the system aggregates the results to form a global verdict for the current iteration. The solution is regarded as valid only if all atomic steps pass verification:
\begin{align}
  V = \bigwedge_{j=1}^{m} v_j.
\end{align}
Simultaneously, the compiled feedback set
\begin{align}
F = \bigcup_{v_j=0} f_j
\end{align}
provides a comprehensive record of all logical flaws, which guides the Augmentation Agent in the subsequent iteration.

Due to the stochastic nature of LLMs, a single ``correct'' verdict may be a false positive, and a single ``wrong'' verdict may be a hallucination. To ensure robustness, the augment-review loop repeats until a confidence threshold is met. We maintain two counters: $n_{\text{corr}}$ (the number of consecutive ``correct'' verdicts) and $n_{\text{wrg}}$ (the cumulative count of ``wrong'' verdicts). The loop terminates when either the solution consistently passes verification, reaching a confidence target: $n_{\text{corr}} \ge N_{\text{corr}}^{(\max)}$, or the solution consistently fails or oscillates sufficient times: $n_{\text{wrg}} \ge N_{\text{wrg}}^{(\max)}$.

\sect{Experimental Setup}
\label{sec:experimental setup}

The CPhO 2025 consists of a theory and an experimental section. The theory section, which is the focus of this work, comprises 7 problems totaling 320 points. Problem 6 is valued at 50 points, while the remaining 6 problems are each 45 points. 

Guided by the CPhO 2025 proposition committee, we created a detailed, step-by-step scoring benchmark. This benchmark was applied uniformly to all evaluations in this paper. Furthermore, owing to the strong fundamental capabilities of state-of-the-art LLMs, their scores are often clustered at the high end. The final margin of improvement, therefore, comes from eliminating the last few mistakes—the most difficult to resolve. For this reason,
we emphasize the \textit{error rate}, as it provides a more sensitive measure of these crucial differences. We define it as the relative difference between the total score and the full score (320 points):
\begin{equation}
    \label{eq:error-rate}
    \text{Error Rate} = \frac{320 - \text{Score}}{320} \times 100\%.
\end{equation}

To evaluate the performance of LOCA, we compare it against several strong baseline methods, ranging from standard prompting strategies to specialized agentic systems:

\begin{itemize}
    \item \textbf{Direct Prompting.} The LLM generates the solution in a single pass without iterative refinement or external augmentation.

    \item \textbf{Chain-of-Thought (CoT)}. We use Zero-Shot-CoT\citep{kojima2022large}, which encourages step-by-step reasoning, and Few-Shot CoT \citep{wei2022chain}, which provides in-context examples of step-by-step reasoning.
    
    \item \textbf{Tree-of-Thoughts (ToT)}. ToT explores a tree of intermediate reasoning steps, allowing the model to self-evaluate and backtrack\citep{yao2023tree}. We configure ToT with a tree depth of $d=4$ and a node size limit of $k=2$.
    
    \item \textbf{Graph-of-Thoughts (GoT)}. GoT is an extension of ToT that organizes the reasoning process into a more flexible graph structure\citep{besta2024graph}.
    
    \item \textbf{Multi-Agent Debate (MAD)}. MAD involves multiple LLM agents that collaboratively propose and critique solutions in a debate format\citep{du2023improving,liang2023encouraging}.  We use 2 agents debating for 3 rounds.
    
    \item \textbf{Self-Refine} \citep{madaan2023self}. A representative iterative method that employs a feedback-refine loop to improve upon the initial raw solution.
    
    \item \textbf{Physics SuperNova (PSN)} \citep{qiu2025physics}. A recently proposed open-source agent system specifically designed to augment LLMs with external tools for solving complex physics problems.
\end{itemize}

\begin{figure*}[t]
	\centering
	\includegraphics[width=0.8\textwidth]{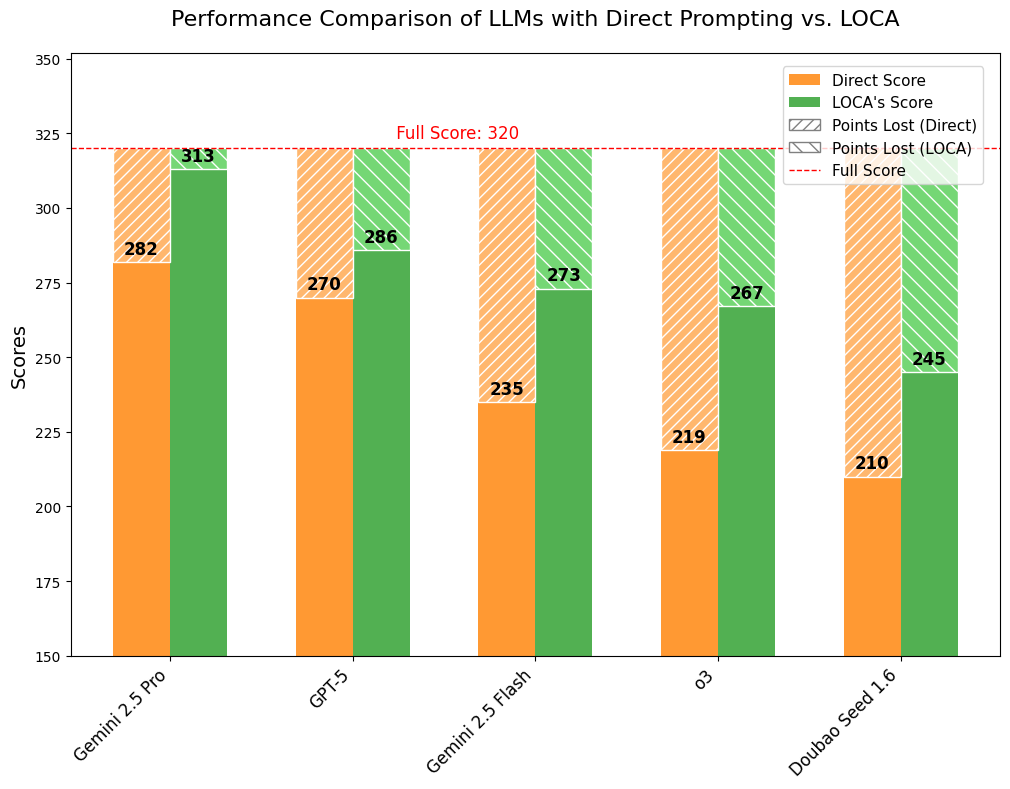} 

    \caption{\textbf{Performance Comparison of LLMs with Direct Prompting vs. LOCA on CPhO 2025}. The chart illustrates the scores of five models (Gemini 2.5 Pro, GPT-5, Gemini 2.5 Flash, o3, and Doubao Seed 1.6) under two different strategies. The height of the solid bars (orange for Direct Prompting, green for LOCA) represents the score achieved by each model. The hatched area above each bar indicates the points lost relative to the full score of 320, which is marked by the red dashed line. The results consistently show that the LOCA method (green) yields higher scores than the direct prompting method (orange) across all tested models.}
	\label{fig:scores_comparison}
\end{figure*}

\sect{Results}

As the CPhO 2025 theory problems incorporate figures, we integrate LOCA with a variety of vision-capable base models, with results presented in Fig.~\ref{fig:scores_comparison}. Notably, all models already achieve superhuman performance (scoring 204) with direct prompting alone, while our LOCA framework yields substantial further improvements. It seems meaningful for LOCA to boost the score of Doubao Seed 1.6 by 35 points. However, even for a highly capable model such as Gemini 2.5 Pro, we still observe a 31-point improvement. This suggests that the gains provided by LOCA are not merely a byproduct of increasing base model capability, but represent a distinct enhancement in reasoning structure.

Notably, LOCA combined with Gemini 2.5 Pro achieves a near-perfect score of \textit{313} out of 320. The remaining errors for this solution are detailed in Supplemental Materials\cite{supp}, providing insights for further refining LOCA to eventually achieve a perfect score.

We next conduct a comparative analysis against various baselines using Gemini 2.5 Pro as the consistent base model. The results, detailed in Tab.~\ref{tab:main_results}, particularly reveal the key advantages of our approach in the high-accuracy regime, where baseline methods begin to plateau.
As shown, LOCA substantially outperforms all other baseline methods, including both general-purpose reasoning frameworks and the domain-specific method PSN. Critically, LOCA's lead is not merely statistical but structural: it correctly solves at least two more sub-problems that remain intractable for other approaches. This ability to bridge the final gap from `high' to `perfect' highlights the significance of our method. Thus, this result establishes a new near-perfect performance benchmark for LLM reasoning on physics problems of Olympiad Level.

Furthermore, we also evaluate LOCA with Gemini 2.5 Pro on the IPhO 2025 to verify the
generalizability of our framework. Given that the competition problems rely heavily on extracting quantitative data from figures, a task distinct from theoretical reasoning, we replace all the original figures with objective textual descriptions as shown in Supplemental Materials\cite{supp}. This ensures that the evaluation isolates the model's theoretical derivation skills.
LOCA again achieves a near-perfect score of 28.6/30 (compared to 26.4/30 for direct prompting under the same conditions). 

While a granular analysis of each component lies outside the primary scope of this main text, we have conducted comprehensive ablation studies to verify their individual contributions, as shown in Supplemental Materials\cite{supp}.

\begin{table}[t]
    \small
    \centering
    \caption{\textbf{Comparison across baseline methods.} Gemini 2.5 Pro is used for all cases, and results are presented as the score of each theory problem, the total score of all 7 theory problems and the error rate defined in Eq.~\ref{eq:error-rate}. Red indicates that some points are deducted for this question, while green indicates that the problem receives full marks. Bold indicates the best performance. LOCA consistently achieves the highest score and the lowest error rate.}
    \vspace{-0.2cm}
    \begin{threeparttable} 
    \resizebox{\columnwidth}{!}{%
    \begin{tabular}{@{}c|ccccccc|c|c @{}}
    \specialrule{.16em}{1.5ex}{.65ex}
         Method  & 1 & 2 & 3 & 4 & 5 & 6 & 7 & Total Score & Error Rate\\
         \specialrule{.07em}{.4ex}{.65ex}
         Human's highest & - & - & - & - & - & - & - & 204 & 36\%\\
         \specialrule{.07em}{.4ex}{.65ex}
         Direct Prompting & \color{ForestGreen}45\color{black} & \color{red}41\color{black} & \color{ForestGreen}45\color{black} & \color{red}33\color{black} & \color{red}39\color{black} & \color{red}39\color{black} & \color{red}40\color{black} & 282 & 12\%\\
         Zero-Shot-CoT & \color{ForestGreen}45\color{black} & \color{red}37\color{black} & \color{ForestGreen}45\color{black} & \color{ForestGreen}45\color{black} & \color{ForestGreen}45\color{black} & \color{red}38\color{black} & \color{red}40\color{black} & 295 & 7.8\%\\
         Few-Shot CoT & \color{ForestGreen}45\color{black} & \color{ForestGreen}45\color{black} & \color{ForestGreen}45\color{black} & \color{red}41\color{black} & \color{ForestGreen}45\color{black} & \color{red}42\color{black} & \color{red}39\color{black} & 302 & 5.6\% \\
         ToT & \color{ForestGreen}45\color{black} & \color{ForestGreen}45\color{black} & \color{ForestGreen}45\color{black} & \color{red}41\color{black} & \color{ForestGreen}45\color{black} & \color{red}40\color{black} & \color{red}39 & 300 & 6.3\%\\
         GoT & \color{ForestGreen}45\color{black} & \color{red}34\color{black} & \color{red}20\color{black} & \color{red}36\color{black} & \color{ForestGreen}45\color{black} & \color{red}39\color{black} & \color{red}39\color{black} & 258 & 19\%\\
         MAD & \color{ForestGreen}45\color{black} & \color{red}33\color{black} & \color{red}42\color{black} & \color{red}43\color{black} & \color{ForestGreen}45\color{black} & \color{red}44\color{black} & \color{red}40\color{black} & 292 & 8.8\%\\
         Self-refine & \color{ForestGreen}45\color{black} & \color{red}43\color{black} & \color{ForestGreen}45\color{black} & \color{red}35\color{black} & \color{red}39\color{black} & \color{red}41\color{black} & \color{red}40\color{black} & 288 & 10\%\\
         PSN & \color{ForestGreen}45\color{black} & \color{red}32\color{black} & \color{red}39\color{black} & \color{red}43\color{black} & \color{ForestGreen}45\color{black} & \color{red}43\color{black} & \color{ForestGreen}45\color{black} & 292 & 8.8\%\\
         \specialrule{.07em}{.4ex}{.65ex}
         LOCA (ours) & \color{ForestGreen}45\color{black} & \color{ForestGreen}45\color{black} & \color{ForestGreen}45\color{black} & \color{ForestGreen}45\color{black} & \color{ForestGreen}45\color{black} & \color{red}43\color{black} & \color{ForestGreen}45\color{black} & \boxed{\textbf{313}} & \textbf{2.2\%}\\
         \specialrule{.16em}{.4ex}{0pt}
    \end{tabular}%
    }
    \end{threeparttable}
    \label{tab:main_results}
\end{table}

\sect{Summary} 
In this work, we introduce LOCA, an AI agent designed to bridge the gap between LLMs and complex physics reasoning. By decomposing solutions into serialized atomic, verifiable steps and employing an iterative augment-review loop, LOCA effectively mitigates the hallucinations inherent in standard solution generation. On the CPhO 2025 theory examination, our method achieved a near-perfect score of 313/320, significantly surpassing both top human performance and strong baselines. These results demonstrate that imposing a rigorous logical architecture unlocks the intrinsic capability of current models to solve problems of exceptional depth. Looking forward, we aim to extend this logic-driven paradigm to broader scientific domains, advancing the development of AI agents capable of serving as trustworthy partners in research and education.

\begin{acknowledgments}
We are grateful to the CPhO 2025 proposition committee for their discussion of the AI-generated results.
This work is supported by the National Natural Science Foundation of China (No. 12325503), and the High-performance Computing Platform of Peking University.
\end{acknowledgments}

% references
\bibliographystyle{utphysMa2}
\bibliography{LOCA}

%%%%%%%%%%%%%%%%%%%%%%%%%%%%%%%%%%%%%%%%%%%%%%%%%
%%%%%%%%%%%%%%%%%%%%%%%%%%%%%%%%%%%%%%%%%%%%%%%%%
\appendix

\end{document}